\documentclass[twoside,11pt]{article}
\usepackage{jmlr2e}

\usepackage{amsmath,amssymb,mathtools}
\usepackage{booktabs}
\usepackage{wrapfig}
\usepackage{subcaption}
\usepackage{bm}
\usepackage{xspace}
\usepackage{url}
\usepackage{cleveref}
\usepackage{xcolor}
\usepackage{cancel}
\usepackage{afterpage}
\usepackage{multirow}

\usepackage{algorithm}
\usepackage{algpseudocode}

\algnewcommand\algorithmicinput{\textbf{Input:}}
\algnewcommand\Input{\item[\algorithmicinput]}
\algnewcommand\algorithmicoutput{\textbf{Output:}}
\algnewcommand\Output{\item[\algorithmicoutput]}

\newcommand{\ours}{\textsc{NeuroSym-BO}\xspace}
\let\cite\citep

\newcommand{\ben}{\begin{enumerate}}
\newcommand{\een}{\end{enumerate}}

\newcommand{\cmt}[1]{}

\usepackage{xspace}

\begin{document}
\firstpageno{1}
\title{Dynamic Bayesian Optimization Framework for Instruction Tuning in Partial Differential Equation Discovery}

\author{
\name Junqi Qu \email jq24b@fsu.edu \\
\addr Department of Computer Science, Florida State University
\AND
\name Yan Zhang \email yzhang18@fsu.edu \\
\addr Department of Computer Science, Florida State University
\AND
\name Shangqian Gao \email sgao@cs.fsu.edu \\
\addr Department of Computer Science, Florida State University
\AND
\name Shibo Li\thanks{Corresponding author.} \email shiboli@cs.fsu.edu \\
\addr Department of Computer Science, Florida State University
}

\begingroup
\renewcommand\thefootnote{\fnsymbol{footnote}} 
\maketitle
\endgroup


\begin{abstract}
	Large Language Models (LLMs) show promise for equation discovery, yet their outputs are highly sensitive to prompt phrasing—a phenomenon we term \textit{instruction brittleness}. Static prompts cannot adapt to the evolving state of a multi-step generation process, causing models to plateau at suboptimal solutions. To address this, we propose \textsc{NeuroSym-BO}, which reframes prompt engineering as a \textit{sequential decision problem}. Our method maintains a discrete library of reasoning strategies and uses Bayesian Optimization to select the optimal instruction at each step based on numerical feedback. Experiments on PDE discovery benchmarks show that adaptive instruction selection significantly outperforms fixed prompts, achieving higher recovery rates with more parsimonious solutions.
\end{abstract}


\section{Introduction}

The automated discovery of physical laws from data is a central challenge in AI for Science~\citep{wang2023scientific,raghu2020survey}. While traditional Symbolic Regression (SR) methods like Genetic Programming (e.g., PySR) are effective, they struggle with combinatorial search spaces and lack semantic priors \cite{cranmer2020discovering}. Large Language Models (LLMs) offer a promising alternative by leveraging pre-trained knowledge of physics and code generation~\citep{chen2021evaluating,wei2022chain}. However, applying LLMs to equation discovery faces a critical barrier: \textit{instruction brittleness}~\citep{sclar2024quantifying,mizrahi2024state}---the phenomenon where small changes in prompt phrasing lead to dramatically different outputs.

Standard frameworks such as LLM-SR \citep{shojaee2024llmsr} and LLM4ED \citep{du2024llm4ed} typically employ fixed prompt templates (e.g., ``Find the equation that fits this data''). This rigid approach mimics a ``static researcher'' unable to adapt their line of questioning. An LLM prompted with ``Find the simplest equation'' might over-regularize and miss important terms; the same model prompted with ``Find the most accurate equation'' might hallucinate spurious terms or fixate on memorized but irrelevant formulas. When faced with complex nonlinear dynamics, static prompts often lead models into local optima from which they cannot escape. This paper makes a key observation: \textbf{the optimal instruction depends on the current state of the generation process}. Early in the search, exploratory prompts (``propose novel functional forms'') are beneficial; later, refinement prompts (``prune redundant terms'') become essential. A fixed prompt cannot capture this state-dependency. 

To overcome this, we introduce \ours, a framework that treats the prompt instruction not as a fixed input, but as a \textit{dynamic hyperparameter} to be optimized during the search. We construct a discrete space of reasoning strategies and employ Bayesian Optimization (BO) to navigate this space, enabling the system to actively switch strategies based on the error profile of current candidates. Our contributions are: \textbf{(1) Dynamic Instruction Tuning}: We formalize prompt engineering in scientific discovery as a discrete Bayesian Optimization problem, enabling adaptive control over the LLM's generation mode. \textbf{(2) Sample Efficiency}: Unlike Reinforcement Learning approaches to prompt optimization, our BO-based method is highly sample-efficient, making it feasible for computationally expensive scientific evaluations. \textbf{(3)Empirical Robustness}: We demonstrate that \textsc{NeuroSym-BO} achieves higher recovery rates on benchmark PDEs, solving cases where static prompting fails.
\section{Background}

\begin{figure*}[!htbp] 
    \centering
    \includegraphics[width=0.9\textwidth]{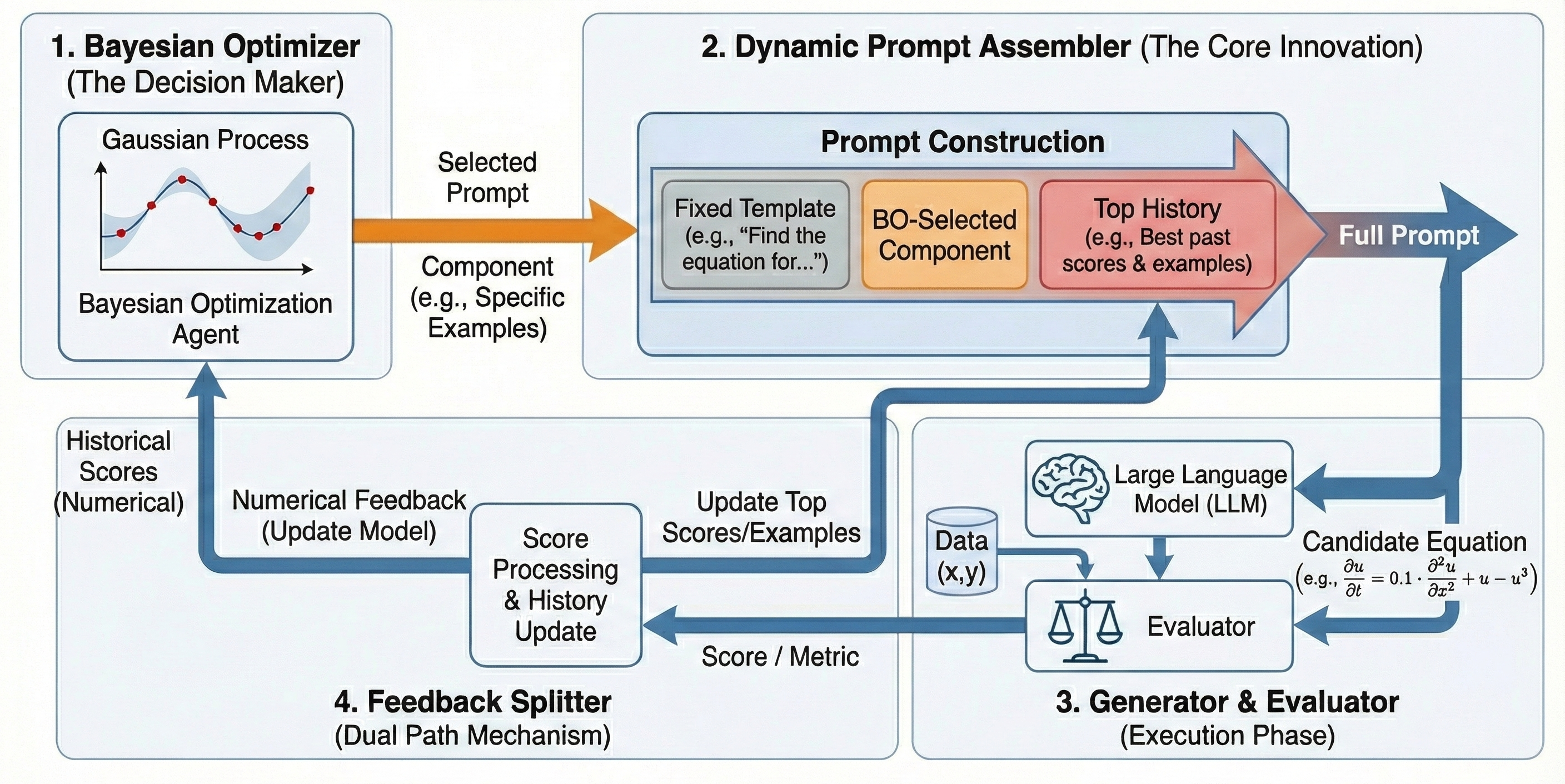} 
    \caption{\small Overview of \textsc{NeuroSym-BO}. A Bayesian Optimizer (1) selects instruction strategies that are assembled into prompts (2) with historical context. The LLM generates candidate equations for evaluation (3), and feedback updates both the history and the optimizer (4), forming a closed-loop system.}
    \label{fig:framework} 

\end{figure*}

\paragraph{PDE Discovery} We consider the task of discovering a governing partial differential equation (PDE) from observational data~\cite{brunton2016sindy,rudy2017pdefind}. Given a spatiotemporal dataset $\mathcal{D} = \{(x_i, t_i, u_i)\}_{i=1}^N$ representing measurements of a physical quantity $u$ at spatial locations $x$ and times $t$, our goal is to recover the symbolic form of the underlying PDE:  $u_t = \mathcal{N}(u, u_x, u_{xx}, \ldots)$ where $\mathcal{N}$ is an unknown nonlinear operator composed of spatial derivatives and algebraic terms. The search space of possible equations is combinatorially large---even with a modest library of operators (addition, multiplication, differentiation), the number of candidate expressions grows exponentially with equation length. This makes exhaustive enumeration intractable and motivates the use of intelligent search strategies.

\paragraph{Bayesian Optimization(BO)} Bayesian Optimization (BO) is a principled framework for optimizing expensive black-box functions \cite{snoek2012practical,jones1998ego,shahriari2016taking}. The key idea is to maintain a probabilistic surrogate model (typically a Gaussian Process)~\citep{rasmussen2006gp} that estimates both the expected value and uncertainty of the objective function at unobserved points. An acquisition function then uses this surrogate to decide which point to evaluate next, balancing the desire to exploit regions with high predicted performance against the need to explore uncertain regions that might contain better solutions.

\section{Methodology}
\label{sec:method}

\ours addresses the challenge of PDE discovery through a three-agent closed-loop architecture (Figure \ref{fig:framework}). The \textbf{Symbolic Proposer} (an LLM) generates candidate equation structures by leveraging its pre-trained knowledge of mathematical forms. The \textbf{Numerical Evaluator} fits coefficients to each candidate and computes a fitness score measuring how well the equation explains the observed data. The \textbf{Prompt Optimizer} (a Bayesian Optimization agent) analyzes feedback from the evaluator and selects the optimal instruction strategy for the next generation round. This closed-loop design allows the system to adaptively refine its search strategy based on accumulated evidence, rather than relying on a fixed generation policy.

\paragraph{Dynamic Prompt Construction} The key innovation of our framework is treating the LLM's instruction as a dynamic, optimizable component. Instead of using a static prompt string throughout the discovery process, we dynamically assemble the prompt $P_t$ at each iteration $t$:
\begin{equation}
    P_t = I_{\text{task}} \oplus \mathcal{H}_{\text{top}} \oplus I_{\text{strategy}}^{(k_t)}
\end{equation}
where $\oplus$ denotes string concatenation. Each component serves a distinct purpose: (1) \textbf{Static Context} ($I_{\text{task}}$): This fixed preamble defines the problem setting, including the state variables (e.g., $u, x, t$), the admissible operator library (e.g., $\{+, \times, \partial_x, \partial_{xx}, \sin, \exp\}$), and output formatting requirements. This ensures the LLM understands the symbolic constraints of the task. (2) \textbf{Dynamic Memory} ($\mathcal{H}_{\text{top}}$): To enable learning from past iterations, we include the top-$N$ best-performing equations discovered so far, along with their fitness scores. This in-context history provides the LLM with implicit ``gradient'' information---by observing which structural patterns achieve high scores, the model can identify promising directions for further exploration. (3) \textbf{Optimizable Instruction} ($I_{\text{strategy}}^{(k_t)}$): This is the core optimizable component. Rather than using a generic directive like ``find the best equation,'' we select a specific reasoning strategy from a pre-constructed \textit{Strategy Bank}~\citep{zhou2023ape,sun2024ease} $\mathcal{B} = \{s_1, \ldots, s_K\}$. We generate $K=100$ diverse instruction variants using a meta-LLM (GPT-4o), covering different aspects of the discovery process. These strategies span a spectrum from \textit{exploration} (``Ignore previous bests. Propose a completely new mathematical structure that has not been tried.'') to \textit{parsimony} (``The current best equations are too complex. Identify and remove terms that contribute least to the fit.''), and include \textit{mutation} directives (``Keep the core structure of the best equation but replace the nonlinear interaction terms with alternatives.'') as well as \textit{refinement} instructions (``The structure looks correct. Focus on adjusting the functional form of individual terms.'').

\paragraph{Bayesian Optimization for Strategy Selection} Selecting the optimal strategy at each iteration is itself an optimization problem. However, evaluating any strategy is expensive: it requires generating candidates from the LLM, fitting their coefficients, and computing residuals against the data. This rules out gradient-based or exhaustive search methods.
We model the mapping from strategy index $k \in \{1, \ldots, K\}$ to the resulting best equation fitness as a black-box function $f: \{1,\ldots,K\} \rightarrow \mathbb{R}$. A Gaussian Process surrogate $\mathcal{GP}$ is fitted to the history of (strategy, fitness) pairs observed so far, providing posterior estimates of the mean $\mu(k)$ and uncertainty $\sigma(k)$ for each strategy. We select the next strategy by maximizing the Expected Improvement (EI) acquisition function:
\begin{equation}
    k_{t+1} = \arg\max_{k} \mathbb{E}_{y \sim \mathcal{GP}}\left[\max(y - y^*, 0)\right]
\end{equation}
where $y^*$ is the best fitness score observed so far. EI quantifies the expected gain from trying strategy $k$: it assigns high values to strategies that either have high predicted performance (exploitation) or high uncertainty (exploration). This principled balance allows the system to efficiently navigate the discrete strategy~\citep{baptista2018bocs} space without exhaustively trying all options.

\paragraph{Numerical Evaluation with Parsimony Penalty} Once the LLM generates candidate symbolic skeletons, we must evaluate their quality. Each candidate is parsed into a symbolic expression, and its free coefficients are optimized using sparse regression (specifically, STRidge) ~\citep{tibshirani1996lasso} to minimize the residual against the observed data $\mathcal{D}$. To prevent overfitting through overly complex equations---a common failure mode in symbolic regression---we design a composite fitness function that balances accuracy against parsimony~\citep{bartlett2024mdl,burlacu2019parsimony}:
\begin{equation}
    S(\hat{u}) = \frac{1 - \lambda \cdot \text{complexity}(\hat{u})}{1 + \text{NRMSE}(\hat{u}, \mathcal{D})}
\end{equation}
where $\text{complexity}(\hat{u})$ counts the number of terms in the equation, $\lambda$ is a penalty coefficient (set to 0.01 in our experiments), and NRMSE $= \sqrt{\text{MSE}/\text{Var}(u)}$ is the normalized root mean square error. This formulation ensures that among equations with similar accuracy, simpler ones receive higher scores. A complete algorithmic description with complexity bounds is given in Appendix~\ref{app:algorithm}.

\begin{figure*}[!htbp]
\centering
\begin{subfigure}{0.33\textwidth}
  \centering
  \includegraphics[width=\linewidth]{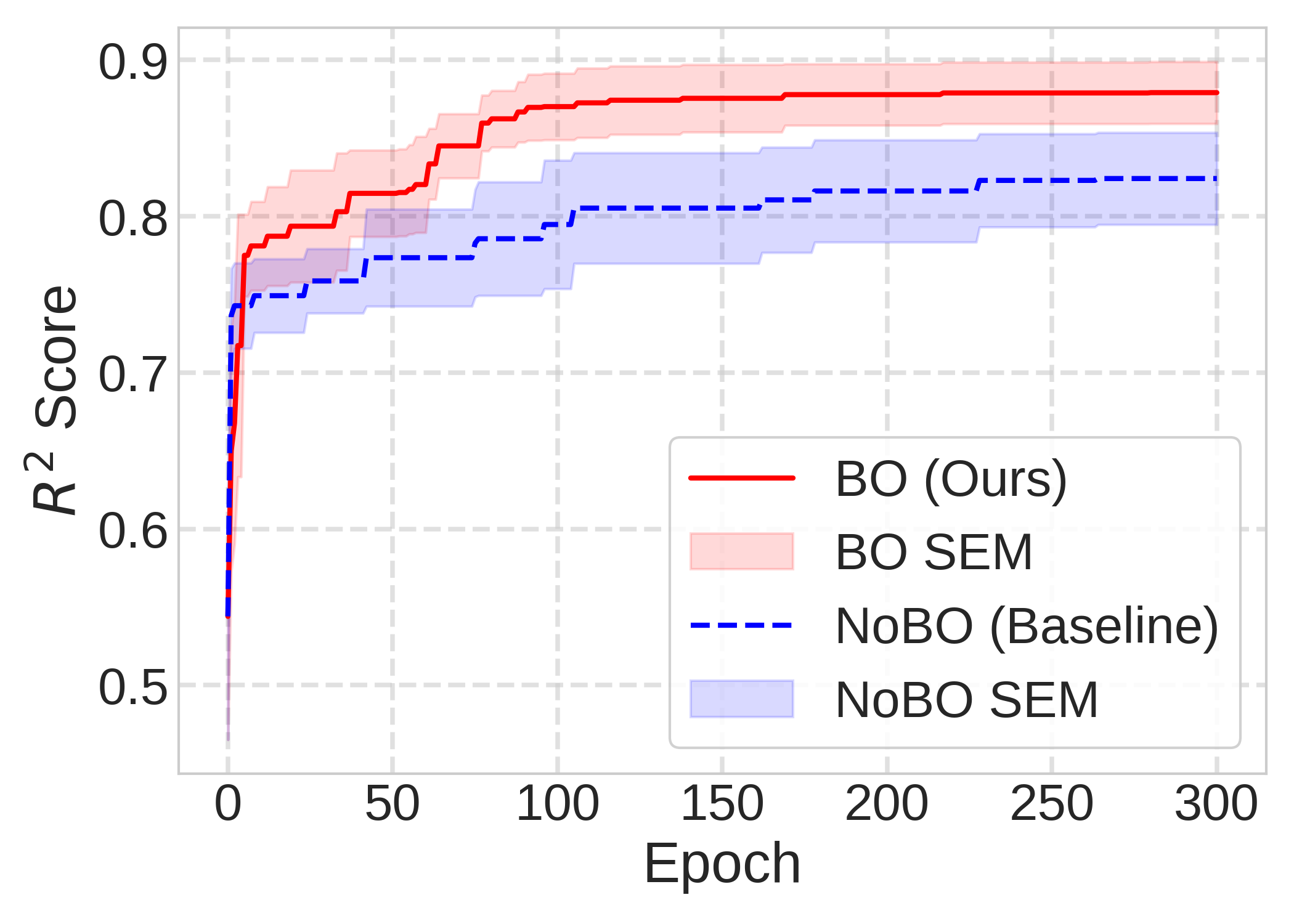}
  \caption{\footnotesize Burgers}
\end{subfigure}\hfill
\begin{subfigure}{0.33\textwidth}
  \centering
  \includegraphics[width=\linewidth]{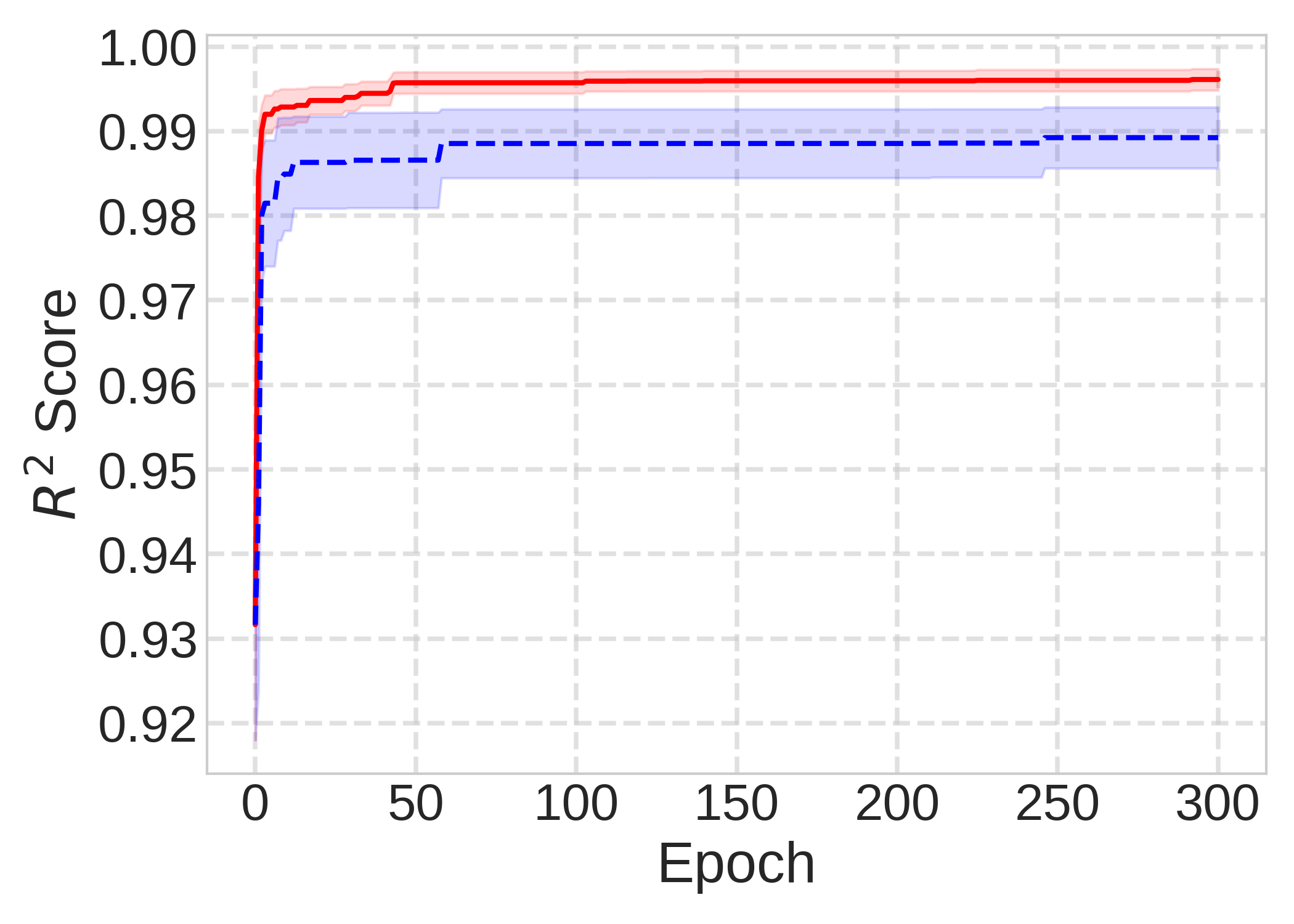}
  \caption{\small Chafee}
\end{subfigure}\hfill
\begin{subfigure}{0.33\textwidth}
  \centering
  \includegraphics[width=\linewidth]{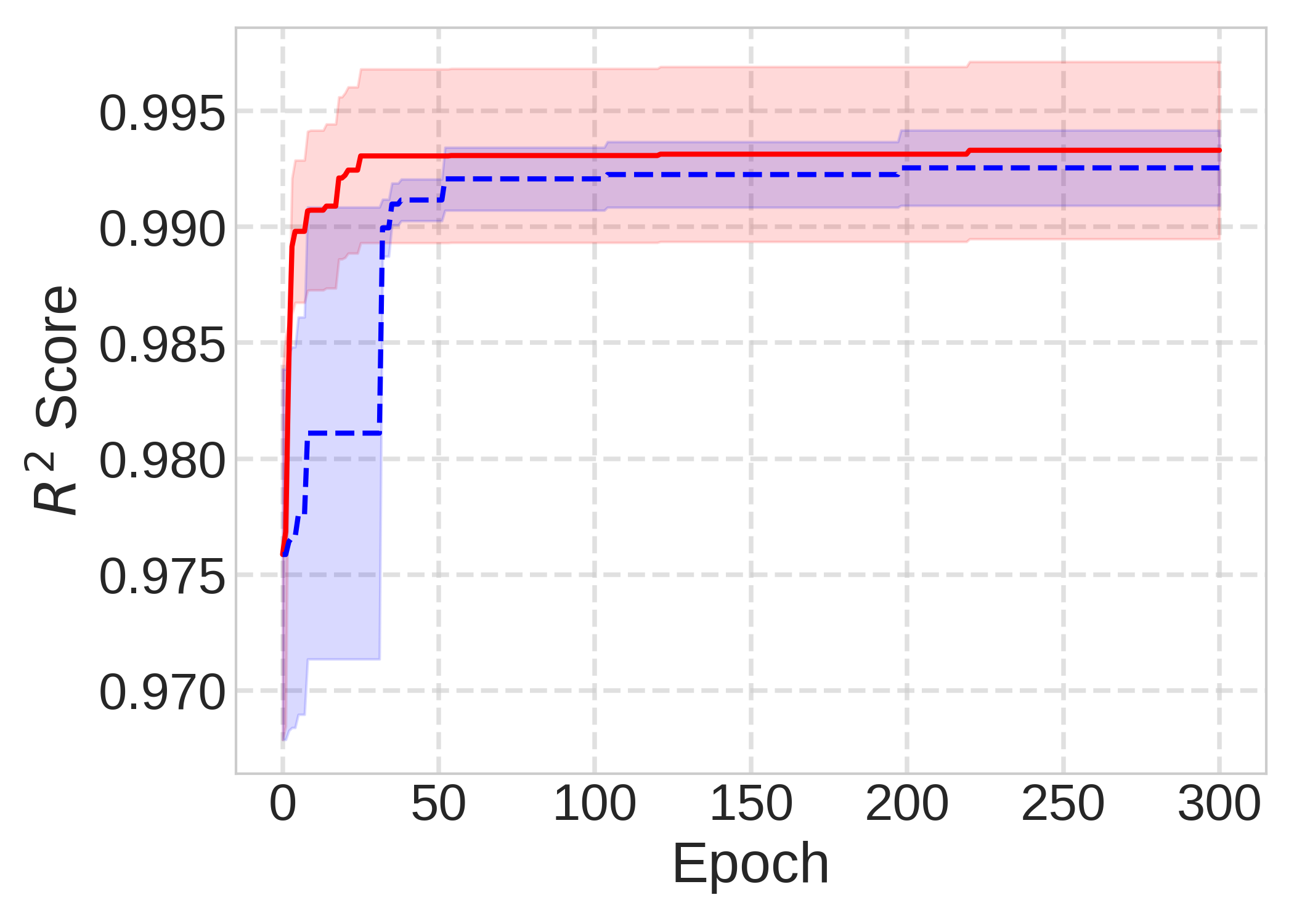}
  \caption{\small Divide}
\end{subfigure}\hfill
\\
\begin{subfigure}{0.33\textwidth}
  \centering
  \includegraphics[width=\linewidth]{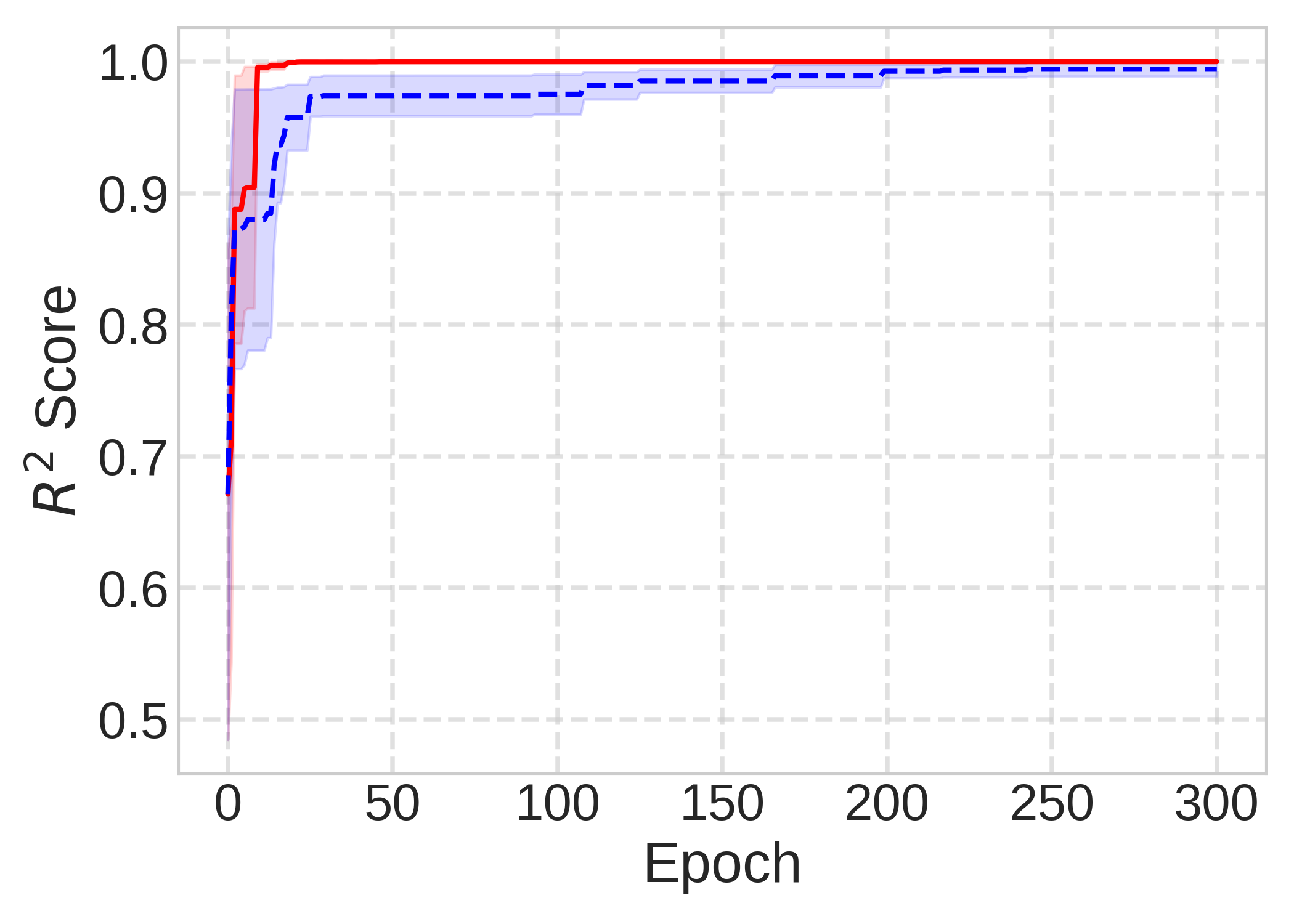}
  \caption{\small Fisher}
\end{subfigure}
\begin{subfigure}{0.33\textwidth}
  \centering
  \includegraphics[width=\linewidth]{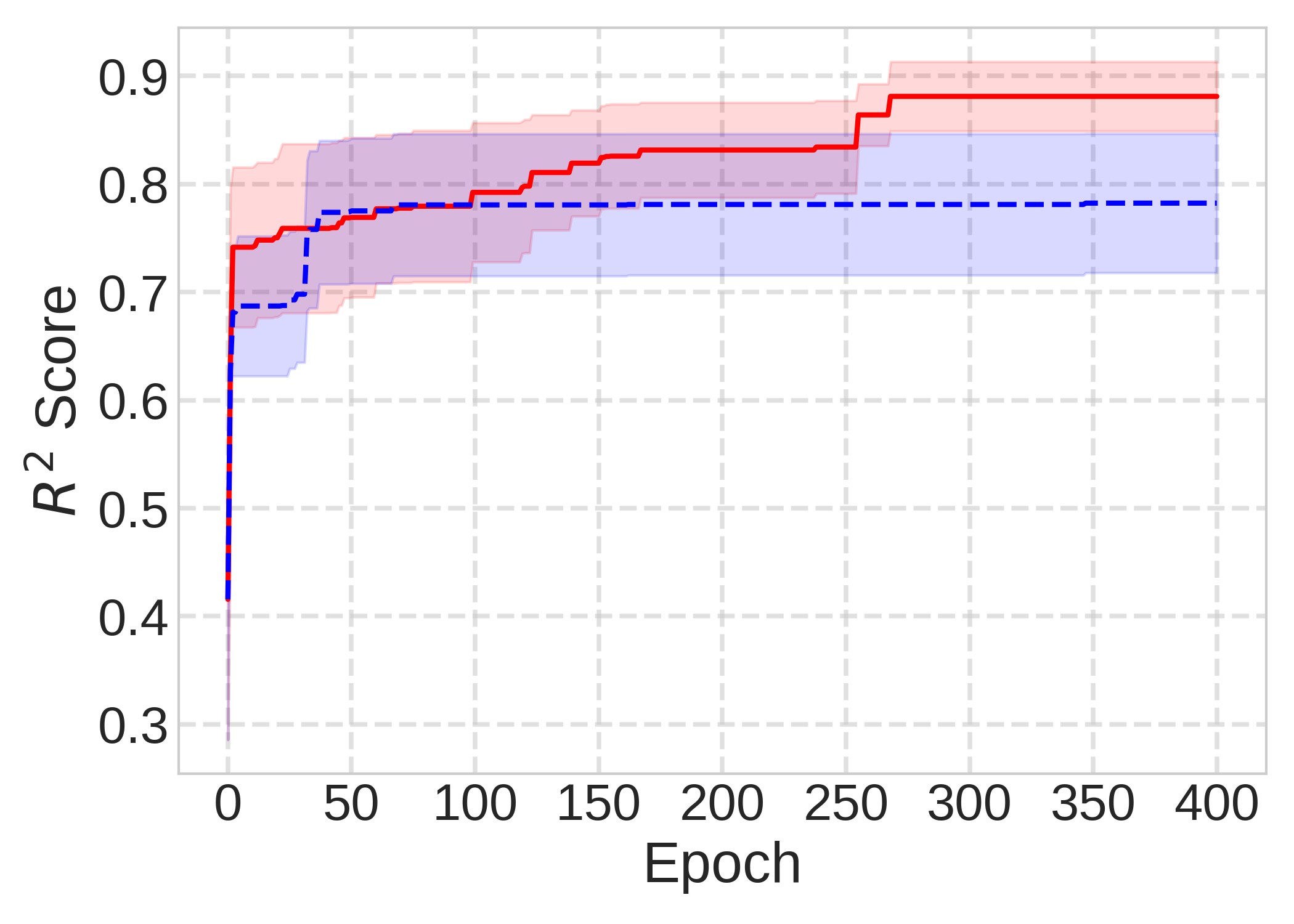}
  \caption{\small Allen-Cahn}
\end{subfigure}\hfill
\caption{\small Optimization trajectories (test $R^2$) across five PDEs: 
red solid: \textsc{NeuroSym-BO}; blue dashed: Fixed Prompt baseline. 
Shaded regions indicate $\pm$SEM over 5 trials. Our method exhibits 
step-wise improvements from adaptive strategy switching, while the 
baseline plateaus early.}
\label{fig:five-in-a-row}
\end{figure*}

\section{Related Work}

Classical equation discovery follows two main approaches. Sparse regression methods like SINDy \cite{brunton2016sindy,messenger2021weak,fasel2022ensemble} and PDE-FIND \cite{rudy2017pdefind} select active terms from predefined libraries via sparsity-promoting optimization. Genetic programming methods—Eureqa \cite{schmidt2009eureqa}, PySR \cite{cranmer2023pysr}, Operon \cite{burlacu2020operon}—evolve expression trees, with physics-informed variants adding domain constraints \cite{zhang2024pigp}. Neural approaches include AI Feynman \cite{udrescu2020aifeynman,udrescu2020aifeynman2,kamienny2022e2e,mundhenk2021seeding,landajuela2022udsr} exploiting physical symmetries, Deep Symbolic Regression \cite{petersen2021dsr} using risk-seeking policy gradients, and transformer models like NeSymReS \cite{biggio2021nesymres} and SymFormer \cite{vastl2022symformer}. KBASS \cite{long2023kbass} combines Bayesian spike-and-slab priors with kernel methods. SRBench \cite{lacava2021srbench} and Feynman equations \cite{udrescu2020aifeynman} provide standard benchmarks. Recent LLM-based methods leverage pre-trained scientific knowledge: LLM-SR \cite{shojaee2024llmsr,boiko2023coscientist,ma2024sga,zhang2024chemllm} uses evolutionary refinement on Python-represented equations, LLM4ED \cite{du2024llm4ed} alternates self-improvement and evolutionary phases, FunSearch \cite{romera2024funsearch} achieved novel mathematical discoveries, LaSR \cite{grayeli2024lasr} builds reusable concept libraries, and ICSR \cite{merler2024icsr} applies in-context learning. However, these methods use static prompts, unable to adapt instructions based on search progress. Prompt optimization methods include APE \cite{zhou2023ape} using bandit selection, OPRO \cite{yang2024opro} with LLMs as meta-optimizers, and evolutionary approaches like EvoPrompt \cite{guo2024evoprompt} and PromptBreeder \cite{fernando2024promptbreeder}. Bayesian prompt optimization has also been explored \cite{sabbatella2024promptbo}. Yet existing methods seek a single optimal prompt, whereas we argue \textit{different prompts are optimal at different stages}. Our work introduces closed-loop instruction optimization via Bayesian Optimization, dynamically selecting strategies based on numerical feedback.

\section{Experiments}
\label{sec:experiments}

We evaluate \ours on five benchmark PDEs: Burgers, Fisher, Chafee-Infante, Divide, and Allen-Cahn (details in Appendix~\ref{app:data}). Both methods use Llama-3.2-3B-Instruct \footnote{\small https://huggingface.co/meta-llama/Meta-Llama-3-8B-Instruct} with identical in-context history (top-5 equations). The \textbf{Fixed Prompt} baseline employs a static instruction throughout, while \textbf{\textsc{NeuroSym-BO}} dynamically selects from the 100-strategy library. We implement the prompt optimizer using BoTorch \footnote{https://botorch.org/} with a standard GP surrogate and EI acquisition. All experiments run for 300 iterations across 5 trials; we report the mean $R^2$ on held-out test data. 
Table~\ref{tab:results} shows \textsc{NeuroSym-BO} consistently outperforms the fixed-prompt baseline, with improvements of 5-11\% on challenging cases (Allen-Cahn, Burgers) and near-perfect recovery on Fisher ($R^2=0.9999$). Figure~\ref{fig:five-in-a-row} visualizes optimization trajectories across all five PDEs. Several patterns emerge: (1) The fixed-prompt baseline (blue, dashed) plateaus early, typically within 50-100 iterations, struggling to escape local optima. (2) \textsc{NeuroSym-BO} (red, solid) exhibits characteristic step-wise improvements, where sudden jumps correspond to the BO agent successfully switching strategies—for instance, transitioning from exploration-focused prompts to simplification directives when the current best equation becomes overly complex. (3) The performance gap widens over time, demonstrating that dynamic instruction selection provides compounding benefits as the search progresses. The shaded regions (±SEM) indicate that our method also achieves lower variance across trials, suggesting more robust convergence.

\begin{table}[!htbp]
\centering
\small
\begin{tabular}{llcccc}
\toprule
\textbf{PDE} & \textbf{Method} & \textbf{Train $R^2$} & \textbf{Test $R^2$} \\
\midrule
\multirow{2}{*}{Allen-Cahn} & Fixed & 0.7968 & 0.7824   \\
 & Ours & \textbf{0.9107} & \textbf{0.8914}   \\
 \midrule
\multirow{2}{*}{Burgers} & Fixed & 0.8102 & 0.8242   \\
 & Ours & \textbf{0.8699} & \textbf{0.8791} \\
\midrule
\multirow{2}{*}{Chafee} & Fixed & 0.9894 & 0.9886 \\
 & Ours & \textbf{0.9951} & \textbf{0.9947}  \\
\midrule
\multirow{2}{*}{Divide} & Fixed & 0.9927 & 0.9922  \\
 & Ours & \textbf{0.9942} & \textbf{0.9941} \\
\midrule
\multirow{2}{*}{Fisher} & Fixed & 0.9952 & 0.9953  \\
 & Ours & \textbf{0.9999} & \textbf{0.9999} \\
\bottomrule
\end{tabular}
\caption{\small Performance comparison across five benchmark PDEs (5-trial average). 
\textsc{NeuroSym-BO} (Ours) consistently outperforms the Fixed Prompt baseline 
on both training and test $R^2$.}
\label{tab:results}

\end{table}

\section{Conclusion}

In this work, we introduced \ours, a closed-loop framework that leverages Bayesian Optimization to dynamically optimize instructions for LLM-based equation discovery. By treating prompt engineering as a search problem over a discrete strategy space, our method effectively overcomes the prompt sensitivity bottleneck inherent in static approaches. Experiments on benchmark PDEs demonstrate that our framework recovers correct symbolic structures with higher success rates while improving sample efficiency. Our findings suggest that combining LLM generative reasoning with numerical optimization feedback is a promising direction for automated scientific discovery.
\section*{Limitations}
While \textsc{NeuroSym-BO} demonstrates promising results, it currently relies on the inherent mathematical capabilities of the backbone LLM (e.g., Llama-3). If the LLM lacks fundamental knowledge of specific mathematical operators, prompt optimization alone cannot solve the problem. Furthermore, our current evaluation focuses on 1D PDEs; scaling to higher-dimensional systems with chaotic behavior remains future work.


\bibliography{reference}



\clearpage
\appendix
\section{Algorithm Overview}
\label{app:algorithm} 

\begin{algorithm}[!htbp]
\caption{\textsc{NeuroSym-BO}: Dynamic Instruction Tuning for PDE Discovery}
\label{alg:neurosymbo}
\begin{algorithmic}[1]
\Require Dataset $\mathcal{D} = \{(x_i, t_i, u_i)\}_{i=1}^N$, Strategy bank $\mathcal{B} = \{s_1, \ldots, s_K\}$, Max iterations $T$, Top-$N$ history size
\Ensure Best discovered equation $\hat{u}^*$

\State Initialize GP surrogate $\mathcal{GP}$, history buffer $\mathcal{H} \leftarrow \emptyset$, observation set $\mathcal{O} \leftarrow \emptyset$
\State $y^* \leftarrow -\infty$ \Comment{Best fitness so far}

\For{$t = 1, \ldots, T$}
    \State \textcolor{blue}{\textit{// Strategy Selection via Bayesian Optimization}}
    \If{$t \leq K_{\text{init}}$} \Comment{Initial exploration phase}
        \State $k_t \leftarrow$ \textsc{RandomSelect}$(\mathcal{B})$
    \Else
        \State Fit $\mathcal{GP}$ to observation set $\mathcal{O}$
        \State $k_t \leftarrow \arg\max_{k} \text{EI}(k; \mathcal{GP}, y^*)$ \Comment{Eq.~2}
    \EndIf
    
    \State \textcolor{blue}{\textit{// Dynamic Prompt Construction}}
    \State $\mathcal{H}_{\text{top}} \leftarrow$ \textsc{TopN}$(\mathcal{H}, N)$ \Comment{Best $N$ equations with scores}
    \State $P_t \leftarrow I_{\text{task}} \oplus \mathcal{H}_{\text{top}} \oplus I_{\text{strategy}}^{(k_t)}$ \Comment{Eq.~1}
    
    \State \textcolor{blue}{\textit{// LLM Generation \& Numerical Evaluation}}
    \State $\{\hat{u}_1, \ldots, \hat{u}_M\} \leftarrow \textsc{LLM}(P_t)$ \Comment{Generate $M$ candidates}
    \For{each candidate $\hat{u}_j$}
        \State $\hat{u}_j \leftarrow$ \textsc{STRidge}$(\hat{u}_j, \mathcal{D})$ \Comment{Fit coefficients}
        \State $S_j \leftarrow \frac{1 - \lambda \cdot \text{complexity}(\hat{u}_j)}{1 + \text{NRMSE}(\hat{u}_j, \mathcal{D})}$ \Comment{Eq.~3}
    \EndFor
    
    \State \textcolor{blue}{\textit{// Feedback \& Model Update}}
    \State $S_t \leftarrow \max_j S_j$; \quad $\hat{u}_t \leftarrow \arg\max_j S_j$
    \State $\mathcal{H} \leftarrow \mathcal{H} \cup \{(\hat{u}_t, S_t)\}$
    \State $\mathcal{O} \leftarrow \mathcal{O} \cup \{(k_t, S_t)\}$ \Comment{Update BO observations}
    \If{$S_t > y^*$}
        \State $y^* \leftarrow S_t$; \quad $\hat{u}^* \leftarrow \hat{u}_t$
    \EndIf
\EndFor

\State \Return $\hat{u}^*$
\end{algorithmic}
\end{algorithm}

Algorithm~\ref{alg:neurosymbo} summarizes the \textsc{NeuroSym-BO} procedure. 
The algorithm operates in three phases per iteration. First, the \textbf{strategy selection} phase (lines 4--10) uses Bayesian Optimization to choose the next instruction strategy $k_t$ from the bank $\mathcal{B}$. During an initial exploration phase ($t \leq K_{\text{init}}$, set to 10 in our experiments), strategies are sampled randomly to build an initial surrogate model. Subsequently, the GP surrogate is fitted to accumulated (strategy, fitness) observations, and the strategy maximizing Expected Improvement is selected.

Second, the \textbf{prompt construction} phase (lines 12--13) assembles the dynamic prompt $P_t$ by concatenating the static task description $I_{\text{task}}$, the top-$N$ best-performing equations from history $\mathcal{H}_{\text{top}}$, and the BO-selected instruction $I_{\text{strategy}}^{(k_t)}$. This provides the LLM with both problem context and implicit feedback about which structural patterns have succeeded.

Third, the \textbf{generation and evaluation} phase (lines 15--20) queries the LLM to produce $M$ candidate equations (we use $M=5$), fits their coefficients via sparse regression, and computes fitness scores balancing accuracy against parsimony. The best candidate updates both the history buffer (for in-context learning) and the BO observation set (for surrogate refinement). This closed-loop design enables the system to progressively refine both \textit{what} equations to propose and \textit{how} to instruct the LLM.

\textbf{Time Complexity.} Let $T$ be the total number of iterations, $K$ the strategy bank size, $M$ the number of LLM-generated candidates per iteration, $N$ the data points in $\mathcal{D}$, and $d$ the number of candidate terms in the operator library.

\begin{itemize}
    \item \textit{Strategy Selection (lines 4--10):} Fitting the GP surrogate to $t$ observations requires $\mathcal{O}(t^3)$ for exact inference due to kernel matrix inversion. Computing EI across $K$ strategies costs $\mathcal{O}(K)$. Per iteration at step $t$: $\mathcal{O}(t^3 + K)$.
    
    \item \textit{Prompt Construction (lines 12--13):} Selecting top-$N$ equations from history $\mathcal{H}$ costs $\mathcal{O}(|\mathcal{H}|)$ with a heap, or $\mathcal{O}(1)$ if maintained incrementally. String concatenation is $\mathcal{O}(L_P)$ where $L_P$ is the prompt length.
    
    \item \textit{LLM Generation (line 15):} Each forward pass through the LLM costs $\mathcal{O}(L_P \cdot d_{\text{model}})$ for context encoding, where $d_{\text{model}}$ is the model dimension. Generating $M$ candidates: $\mathcal{O}(M \cdot L_P \cdot d_{\text{model}})$.
    
    \item \textit{Coefficient Fitting (line 17):} STRidge sparse regression for each candidate costs $\mathcal{O}(N \cdot d^2)$ per iteration of the thresholding loop. With $M$ candidates: $\mathcal{O}(M \cdot N \cdot d^2)$.
    
    \item \textit{Fitness Evaluation (line 18):} Computing NRMSE requires $\mathcal{O}(N)$ operations; complexity counting is $\mathcal{O}(d)$. Total: $\mathcal{O}(M \cdot N)$.
\end{itemize}

The total time complexity over $T$ iterations is $\mathcal{O}\left(\sum_{t=1}^{T} t^3 + T \cdot \left(K + M \cdot (C_{\text{LLM}} + N \cdot d^2)\right)\right) = \mathcal{O}(T^4 + T \cdot M \cdot C_{\text{LLM}})$ where $C_{\text{LLM}}$ denotes the LLM inference cost. In practice, the GP cubic term can be mitigated using sparse GP approximations~\cite{quinonero2005unifying}, reducing it to $\mathcal{O}(T \cdot m^2)$ with $m \ll T$ inducing points. The dominant cost is typically LLM inference.

\textbf{Space Complexity.} The algorithm maintains: (1) the history buffer $\mathcal{H}$ storing $\mathcal{O}(T)$ equations, (2) the GP observation set $\mathcal{O}$ of size $\mathcal{O}(T)$, and (3) the kernel matrix of size $\mathcal{O}(T^2)$. Total space complexity is $\mathcal{O}(T^2 + T \cdot L_{\text{eq}})$, where $L_{\text{eq}}$ is the average equation length.

\textbf{Practical Considerations.} In our experiments with $T=300$, $K=100$, $M=5$, and Llama-3.2-3B, each iteration completes in approximately 15--30 seconds on a single A100 GPU, with LLM inference accounting for $\sim$80\% of the runtime. The GP overhead remains negligible ($<$1\%) due to the moderate number of iterations.

\section{Dataset and Equation Details}
\label{app:data}

To ensure rigorous and reproducible evaluation, we utilize a combination of established benchmark datasets and self-generated simulations that span a diverse range of physical phenomena. For the Burgers, Fisher, Chafee-Infante, and PDE\_divide equations, we employ standard datasets from the \textbf{LLM4ED} benchmark suite~\cite{du2024llm4ed}, which has been widely adopted for evaluating symbolic equation discovery methods. Additionally, we generate a synthetic dataset for the Allen-Cahn equation to test our framework's capability on phase-separation dynamics characterized by sharp interfaces and bistable nonlinearities. Together, these five PDEs represent a comprehensive testbed covering fluid dynamics, population biology, pattern formation, and materials science.

\subsection{Benchmarks from LLM4ED}

The following equations are adopted from the LLM4ED repository~\cite{du2024llm4ed}. We provide their canonical forms, physical interpretations, and simulation configurations below.

\paragraph{Burgers' Equation.} 
Burgers' equation is a fundamental partial differential equation in fluid mechanics that serves as a simplified model for shock wave formation, turbulence, and nonlinear acoustics~\cite{burgers1948mathematical,whitham2011linear}. First introduced by Bateman~\cite{bateman1915some} and later extensively studied by Burgers~\cite{burgers1948mathematical}, it combines nonlinear convection with diffusive dissipation, making it an ideal testbed for symbolic discovery methods. The viscous form is given by:
\begin{equation}
    u_t + u u_x = 0.1 u_{xx}
\end{equation}
where the left-hand side represents nonlinear advection (the term $u u_x$ causes wave steepening) and the right-hand side represents viscous diffusion with coefficient $\nu = 0.1$. The interplay between these terms leads to the formation of shock-like structures that eventually smooth out due to diffusion. The celebrated Cole-Hopf transformation~\cite{cole1951quasi,hopf1950partial} provides an analytical framework for understanding these solutions.

\textit{Setup:} The simulation is performed on a spatial domain $x \in [-8, 8]$ over a time interval $t \in [0, 10]$. The data is discretized on a uniform grid of size $256 \times 201$ (spatial $\times$ temporal points). The initial condition consists of a smooth profile that evolves into a traveling shock wave, providing rich dynamics for equation discovery.

\paragraph{Fisher's Equation (Fisher-KPP).} 
Fisher's equation, also known as the Fisher-Kolmogorov-Petrovsky-Piskunov (Fisher-KPP) equation, is a classical reaction-diffusion model originally proposed by Fisher~\cite{fisher1937wave} and independently by Kolmogorov, Petrovskii, and Piskunov~\cite{kolmogorov1937study} to describe the spatial spread of advantageous genes in a population. It has since found applications in ecology, epidemiology, and combustion theory~\cite{murray1989mathematical}. The equation takes the form:
\begin{equation}
    u_t = u_{xx} + u(1-u)
\end{equation}
where $u_{xx}$ represents spatial diffusion and $u(1-u)$ is a logistic growth term that drives the population toward carrying capacity. This equation admits traveling wave solutions~\cite{ablowitz1979explicit} that propagate at a minimum speed determined by the linearization at the unstable equilibrium $u=0$.

\textit{Setup:} The simulation domain is $x \in [-1, 1]$ with time interval $t \in [0, 1]$. The data is discretized on a $200 \times 100$ grid. The initial condition is chosen to exhibit front propagation behavior, testing the method's ability to recover both diffusive and reactive terms simultaneously.

\paragraph{Chafee-Infante Equation.} 
The Chafee-Infante equation is a reaction-diffusion PDE that arises in the study of phase transitions and pattern formation~\cite{chafee1974bifurcation}. It is closely related to the Allen-Cahn equation and exhibits bistable dynamics with two stable equilibria at $u = \pm 1$. The geometric theory of such semilinear parabolic equations has been extensively developed~\cite{henry1981geometric}. The equation is given by:
\begin{equation}
    u_t - u_{xx} = u - u^3
\end{equation}
The cubic nonlinearity $u - u^3$ creates a double-well potential structure, leading to the formation of domain walls (interfaces) separating regions of different phases. This equation has been extensively studied in the context of chaotic attractors and infinite-dimensional dynamical systems.

\textit{Setup:} The computational domain is $x \in [0, 3]$ with time interval $t \in [0, 0.5]$. The data is discretized on a $301 \times 200$ grid. The relatively short time interval captures the initial transient dynamics and interface formation, providing a challenging test case for symbolic recovery of the cubic reaction term.

\paragraph{PDE\_divide (Synthetic Division Test).} 
This synthetic benchmark is specifically designed to evaluate the capability of symbolic discovery methods to recover rational terms involving division operators (e.g., $1/x$ or $u/x$). Such terms pose significant challenges for traditional genetic programming approaches, which often struggle with protected division operations and singularity handling. The equation is:
\begin{equation}
    u_t = 0.25 u_{xx} - \frac{u_x}{x}
\end{equation}
This PDE can be interpreted as a diffusion equation with a spatially-varying advection term that becomes singular at $x=0$. The coefficient $-1/x$ in front of $u_x$ represents a radially-dependent drift in cylindrical or spherical coordinates.

\textit{Setup:} The domain is $x \in [1, 2]$ with time interval $t \in [0, 1]$. Importantly, the spatial domain is chosen to exclude $x=0$, thereby avoiding the singularity while still requiring the discovery method to identify the $1/x$ dependence. The data is discretized on a $100 \times 251$ grid. This benchmark specifically tests whether \textsc{NeuroSym-BO} can discover non-polynomial functional forms that are difficult for standard symbolic regression methods.

\subsection{Allen-Cahn Equation}

To further evaluate our framework on physically meaningful problems beyond the LLM4ED suite, we generate a synthetic dataset for the Allen-Cahn equation. This equation is a fundamental model in materials science and mathematical physics, describing phase separation phenomena in binary alloys, order-disorder transitions, and interface motion~\cite{allen1979microscopic}. The governing equation is:
\begin{equation}
    u_t = 0.1 u_{xx} + 5.0 (u - u^3)
\end{equation}
Here, the diffusion coefficient $D = 0.1$ controls the interface width, while the reaction rate $R = 5.0$ determines the strength of the bistable nonlinearity. The term $(u - u^3)$ derives from the derivative of a double-well free energy potential $F(u) = \frac{1}{4}(1-u^2)^2$, originally introduced by Cahn and Hilliard~\cite{cahn1958free} in their seminal work on phase separation, with stable equilibria at $u = \pm 1$ representing two distinct phases.

\paragraph{Physical Significance.}
The Allen-Cahn equation exhibits rich dynamics including phase coarsening (where smaller domains shrink and larger ones grow), interface annihilation, and curvature-driven motion~\cite{bates1997spectral}. The sharp interfaces between phases make this equation particularly challenging for numerical methods~\cite{du2019maximum} and, consequently, for data-driven discovery approaches that must accurately capture both the smooth bulk dynamics and the steep gradients at interfaces.

\paragraph{Numerical Generation.}
We generate ground truth data using high-accuracy numerical methods to ensure reliable training and evaluation. Specifically, we employ a pseudo-spectral method~\cite{trefethen2000spectral,boyd2001chebyshev} for spatial discretization, which provides exponential convergence for smooth solutions and accurately resolves the steep gradients at phase boundaries. Time integration is performed using an adaptive Runge-Kutta scheme (RK45)~\cite{dormand1980family} that automatically adjusts step sizes to maintain accuracy.

\begin{itemize}
    \item \textbf{Parameters:} The diffusion coefficient is set to $D=0.1$ and the reaction rate to $R=5.0$. These values are chosen to produce well-separated timescales between diffusion and reaction, resulting in sharp but resolvable interfaces.
    
    \item \textbf{Domain \& Grid:} The simulation is performed on a one-dimensional spatial domain $x \in [-10, 10]$ (total length $L=20$) over a time interval $t \in [0, 10]$. We discretize the system on a uniform grid of size $N_x \times N_t = 256 \times 201$, providing sufficient resolution to capture interface dynamics while maintaining computational efficiency.
    
    \item \textbf{Initial Condition:} To generate complex, physically realistic dynamics with multiple interacting interfaces, we initialize the system with a composite trigonometric function perturbed by random noise:
    \begin{equation}
        u(x,0) = \sin\left(\frac{2\pi x}{L}\right) + 0.5\cos\left(\frac{4\pi x}{L}\right) + \epsilon
    \end{equation}
    where $L=20$ is the domain length and $\epsilon \sim \mathcal{U}(0, 0.2)$ represents uniformly distributed initialization noise. This initial condition creates multiple zero-crossings that evolve into sharp interfaces, testing the discovery method's ability to handle multi-scale dynamics.
    
    \item \textbf{Boundary Conditions:} We impose periodic boundary conditions, which are naturally handled by the spectral method and ensure that no artificial boundary effects contaminate the interior dynamics.
    
    \item \textbf{Solver Details:} Spatial derivatives ($u_{xx}$) are computed in the frequency domain using the Fast Fourier Transform (FFT), which provides spectral accuracy for periodic problems~\cite{trefethen2000spectral}. The semi-discrete system of ODEs is integrated using \texttt{scipy.integrate.solve\_ivp}\footnote{https://docs.scipy.org/doc/scipy/reference/integrate.html}~\cite{virtanen2020scipy} with the RK45 method. Both relative and absolute tolerances are set to $10^{-8}$ to ensure high-fidelity data generation. The solver adaptively refines the time step during periods of rapid interface motion.
    
    \item \textbf{Data Extraction:} After simulation, we uniformly sample the solution at $201$ time points and store the full spatial field at each time, resulting in a data tensor of shape $(256, 201)$. We further split this data into training (80\%) and testing (20\%) sets along the temporal axis for evaluation.
\end{itemize}

\paragraph{Rationale for Inclusion.}
The Allen-Cahn equation complements the LLM4ED benchmarks by introducing several additional challenges: (1) stronger nonlinearity with the coefficient $R=5.0$ compared to the Chafee-Infante equation, (2) longer time evolution allowing observation of phase coarsening dynamics, and (3) a different balance between diffusion and reaction terms. Successfully discovering this equation demonstrates that \textsc{NeuroSym-BO} generalizes beyond the specific parameter regimes present in existing benchmarks.


\end{document}